# Performance of a domain-specific large language model in answering patient questions in psychiatry

Alexander J. Hish, MD[a], Arjun Nagendran, PhD[a,b], Scott N. Compton, PhD[a,c]

[a]Pritzker Department of Psychiatry and Behavioral Health, Lurie Children's Hospital of Chicago, Chicago, IL, USA

[b]Relativ.ai, USA

[c]Northwestern University Feinberg School of Medicine, Chicago, IL, USA

## 1. Introduction

Large language models (LLMs) have demonstrated promise in generating healthcare-related responses, but with important limitations. Studies in general medicine and psychiatry show that LLMs (including the current most widely-used LLM, ChatGPT) can produce fluent, informative answers, but often lack nuance, omit safety considerations, or introduce hallucinations and overgeneralizations that may mislead patients (Asgari et al., 2025; Luykx et al., 2023; Peters & Chin-Yee, 2025; Senturk & Koparal, 2026). Systematic reviews highlight that methods used to evaluate LLM responses in medicine are heterogeneous and frequently rely on non-clinician raters, raising concerns about overstated safety (Tam et al., 2024; Wei et al., 2024).

At the same time, domain-specific fine-tuning and restricted training corpora can improve fidelity and reduce hallucinations (Chen et al., 2024; Van Veen et al., 2023). One example of an LLM fine-tuned for medical decision-making is OpenEvidence, a model trained solely on peer-reviewed medical literature (Wu & Casauay, 2024). One study showed that OpenEvidence can produce evidence-based recommendations in the primary care setting, aligning with physician-generated treatment plans (Hurt et al., 2025). However, this platform is not designed for patients, and demonstrates shortcomings such as sociodemographic biases when used for mental health purposes (Jia et al., 2026).

While a prior study has evaluated how general purpose LLMs compare to psychiatrists in answering patient questions about antidepressants, to our knowledge, no studies thus far have examined whether a domain-specific LLM can answer psychiatric medication-related questions for a patient audience (Senturk & Koparal, 2026). This gap is particularly relevant in psychiatry, where public information sources on psychiatric medications are themselves inconsistent and sometimes inaccurate (Demasi & Gøtzsche, 2020). Addressing whether a targeted model can provide accurate, patient-centered information—while avoiding the errors of public chatbots—would clarify the role of restricted-corpus LLMs in safe clinical communication.

A model like this could improve the efficiency of psychiatry visits, while also improving patient outcomes. In community settings, psychiatrists may spend as little as 15-20 minutes per follow-up visit, much of which may be devoted to addressing questions about medications (Torrey et al., 2017). This time constraint, combined with rising caseloads and workforce shortages, limits opportunities for thorough patient education, leading many patients to seek information from unverified online sources that may amplify fears or misconceptions about psychiatric treatment (Schumacher et al., 2025). Medication adherence remains one of the greatest challenges in clinical psychiatry, with nearly half of patients with major psychiatric disorders demonstrating partial or full non-adherence (Semahegn et al., 2020). A substantial proportion of adherence issues arise from limited understanding of medications, inadequate time for questions during brief clinical encounters, and weak therapeutic alliance (Laranjeira et al., 2023). Poor adherence contributes to symptom relapse, hospitalization, and other adverse outcomes, while increasing healthcare costs (Bhattacharyya, 2023).

By providing accurate, personalized, and on-demand explanations about medication purpose, side effects, and related topics, an LLM could answer many of the common questions that currently consume valuable visit time. This between-visit engagement would allow psychiatrists to focus on individualized decision-making and therapeutic rapport while ensuring that patients receive evidence-based information. Moreover, by improving understanding and addressing misconceptions, a reliable LLM-based educational tool could promote sustained adherence to medication (Loots et al., 2021; Moulaei et al., 2025).

To this end, this study aims to evaluate whether an LLM trained exclusively on patient education resources can answer common questions about psychiatric medications, while avoiding the inaccuracies and safety issues commonly observed in LLM chatbots. The psychiatric medication escitalopram was chosen as a case study.

## 2. Methods

We developed an LLM (codenamed "MIND") fine-tuned for clinical alignment and fidelity, and trained on patient education resources curated from some of the most authoritative English-language psychiatric and medical organizations. We compared the responses of MIND, ChatGPT, and OpenEvidence to common patient questions about the psychiatric medication escitalopram.

### *2.1 Model Development*

Development Principles: MIND was developed using the following three guiding principles:

- Clinical Alignment: Prioritizing safety and adherence to psychiatric best practices through retrieval-grounded responses and provider oversight.
- Data Privacy: Processing only deidentified, publicly available, or provider-generated materials within secure, air-gapped environments.
- Responsible AI Deployment: Ensuring transparency, reproducibility, and continuous monitoring through auditable logs and modular architecture.

Preprocessing: Input materials were standardized through a multistage preprocessing pipeline implemented in Python using Pandas, NumPy, and regular expressions. Document files were processed via Tesseract OCR to extract and normalize text content. All metadata and identifiable text were removed prior to ingestion. The resulting corpus was indexed into a vectorized knowledge store for efficient semantic retrieval.

Model Training & Fine-Tuning: The system employs locally hosted foundation models in the 7–20 billion parameter range (e.g., open-weight LLaMA-family models), accessed securely through LM Studio or Ollama interfaces. Rather than full-parameter fine-tuning, models were optimized through domain adaptation using in-context learning. Provider-approved examples of clinically validated responses were embedded directly in the context window during inference. This approach maintained data security by avoiding gradient updates on clinical data while achieving task-specific alignment.

Model prompts were constructed using LangChain and templated with Jinja2 to enforce structured, clinically interpretable output formats. The prompt templates guided the model to include:

- Evidence citation or reference when available.
- Clear indication of when provider verification is recommended.
- Warnings against providing diagnostic or prescriptive advice beyond patient education scope.

Retrieval Augmented Generation Pipeline: The RAG subsystem combined semantic retrieval with LLM-based reasoning to improve factuality and reduce hallucination. When a patient query is received:

- The query is embedded using a local embedding model and matched against the vector database of provider-approved references.
- The top-ranked passages are retrieved and appended to the model prompt.
- The LLM generates a response conditioned on both the query and retrieved evidence.

This hybrid approach ensures that generated outputs remain grounded in verifiable clinical knowledge while benefiting from the language fluency and contextual adaptability of large generative models.

Infrastructure & Security: All computation occurs on air-gapped, high-performance servers equipped with GPU acceleration (e.g., NVIDIA RTX 5090, CUDA/cuDNN). The processing environment enforces complete network isolation to prevent data leakage. Controlled remote access for system administrators and authorized providers is managed via ephemeral port forwarding and dynamic IP routing through ngrok. Optional integration with Groq Cloud enables accelerated inference with Zero Data Retention (ZDR), ensuring that no user or query data are persisted outside the local environment.

### *2.2 Training Corpus*

All data used in model development and retrieval were publicly available, ensuring compliance with data privacy and regulatory standards. Patient education resources included escitalopram medication information from seven authoritative sources: American Academy of Child and Adolescent Psychiatry; American Academy of Pediatrics; American Psychiatric Association; Food and Drug Administration; National Alliance on Mental Illness; National Library of Medicine (managed by the National Institutes of Health); and UpToDate (a highly peer-reviewed online medical resource). (Resource URLs are included in the supplement.) These data sources encompass structured FAQ datasets, peer-reviewed psychiatric medication guides, and standard clinical reference materials.

### *2.3 Model Queries*

Each model was queried with 50 questions in English. (Questions are listed in the supplement.) Questions were designed by the psychiatrist on our team to be frequently asked questions about various aspects of escitalopram treatment, including mechanisms of action, indications, dosing, common and rare side effects, interactions, and special cases, encompassing all aspects of a standard informed consent discussion (Joint Commission on Accreditation of Healthcare Organizations, 2016). A few questions were vignette-style and included contextual details about a hypothetical patient case. Models were instruction-tuned with the instruction: "Provide answers to the following questions in less than 100 words, in a style that would be readable and understandable by the average layperson."

### *2.4. Model Response Ratings*

Model responses were rated according to two methods. In the first method, LLM responses (from MIND, ChatGPT, and OpenEvidence) were graded according to a rubric measuring their accuracy, clarity, completeness, nuance, safety, and referral appropriateness on a 1-5 Likert scale (1=low, 5=high). This involved an automated analysis of responses comparing each response to the items as specified on our rubric, using reinforcement learning from human feedback and few-shot learning. Rubric constructs were defined as follows:

- Accuracy: The response accurately reflects the evidence base.
- Clarity: The response is easily readable and understandable by a layperson.
- Completeness: The response completely answers the question.
- Nuance: The response acknowledges uncertainty and the importance of context.
- Safety: The response does not provide harmful recommendations.
- Referral Appropriateness: The response appropriately refers to a physician when indicated (and does not when not indicated).

Non-parametric tests (Kruskal-Wallis omnibus tests and Dwass-Steel-Critchlow-Fligner pairwise comparisons) were used to compare mean scores between the three LLMs.

Flesch-Kincaid readability scores (as calculated by Google Gemini) were used as a secondary measure to assess readability of LLM responses.

In the second method, following study designs from recent studies evaluating LLM responses to patient questions, responses from MIND and ChatGPT were rated by psychiatrists licensed by the American Board of Psychiatry and Neurology (Luykx et al., 2023; Peters & Chin-Yee, 2025; Ye et al., 2024). OpenEvidence was not included in psychiatrist ratings because it is not available to the public, and to ease the survey burden. Psychiatrists were recruited through email flyers sent to academic psychiatry departments. Participants were screened to recruit participants with a range of clinical experience to enhance generalizability. Participation was voluntary, all participants completed informed consent procedures, and participants were compensated a $50 gift card after survey completion. This study was determined to be Non-Human Subjects Research by the Lurie Children's Institutional Review Board (Study ID #00000824).

Psychiatrists rated LLM responses according to their accuracy, completeness, and safety, on a 1-3 scale (1=disagree, 2=unsure, 3=agree). For the purposes of psychiatrist review, these constructs were defined as follows:

- Accuracy: The response accurately reflects the evidence base.
- Completeness: The response completely answers the question, for example, by acknowledging uncertainty and relevant context.
- Safety: The response does not provide harmful recommendations.

Raters were also asked to decide whether the question needed to be discussed with a doctor (to assess referral appropriateness), and to select their preferred response between the two. To avoid overwhelming psychiatrist raters with too many constructs, we removed the "clarity" construct from this survey (instead using a secondary measure of readability, Flesch-Kincaid scores, as above), and collapsed the "nuance" construct into the "completeness" construct. Raters were blinded to the source of each response, but were informed that all responses were generated by LLMs.

Kendall's W was used to evaluate inter-rater reliability. Non-parametric tests were used to compare domain scores of accuracy, completeness, and safety between the two LLMs. Specifically, Binomial Test was used to compare preferences, and Mann-Whitney U test was used to compare mean domain scores. To compare referral appropriateness, the psychiatrist on our team coded each LLM response as 1 (referral was made) or 2 (referral was not made), based on whether the response included text such as "ask your doctor", "talk to your doctor", or "your doctor will decide". Then, Spearman's Rho was used to evaluate correlation with psychiatrist ratings.

All statistical tests, unless otherwise specified, were conducted in Jamovi desktop version (The Jamovi Project, 2025).

## 3. Results

### *3.1 Sample and Response Characteristics*

MIND was able to provide responses to 38 of the 50 questions; for the remaining 12 questions, it indicated that an answer was not available. ChatGPT and OpenEvidence were able to provide responses to all 50 questions. (Responses are included in the supplement.) Each model produced responses adhering to the instruction limit of <100 words (except in the case of 6 responses for MIND; and for OpenEvidence, which was unable to be instruction-tuned), with average word count of 83.8 (SD=19.4) for MIND, 45.2 (SD=6.4) for ChatGPT, and 84.7 (SD=27.7) for OpenEvidence. For the 38 questions that MIND was able to answer, responses from each LLM were included in the analysis (38 responses x 3 models = 114 responses).

***3.2 Construct Scores, Rated by Rubric***

Construct scores, as rated by rubric, are summarized in Table 1. Significant differences were observed in all constructs, with MIND rated highest in all domains: accuracy ($X^2$=68.9, $p<0.001$, $\varepsilon^2$=0.610), clarity ($X^2$=43.3, $p<0.001$, $\varepsilon^2$=0.384), completeness ($X^2$=60.0, $p<0.001$, $\varepsilon^2$=0.531), nuance ($X^2$=42.0, $p<0.001$, $\varepsilon^2$=0.371), safety ($X^2$=37.9, $p<0.001$, $\varepsilon^2$=0.336), and referral appropriateness ($X^2$=19.2, $p<0.001$, $\varepsilon^2$=0.170).

***3.3 Readability, Rated by Flesch-Kincaid Score***

Flesch-Kincaid readability scores differed significantly between LLMs (ANOVA $p<0.001$), with ChatGPT rated first (mean=60.5, SD=8.9), MIND rated second (mean=47.8, SD=10.1), and OpenEvidence rated third (mean=43.0, SD=10.7). This corresponded to average grade levels of 8.8, 11.2, and 11.8, respectively.

***3.4 Psychiatrist Demographics***

N=10 psychiatrists rated LLM responses. Of these, 4 were considered early career (0-4 years of post-board licensure experience), 4 were middle career (5-9 years), and 2 were late career (≥10 years), with a range of <1 to 25 years and average 7.8 years of experience. 5 held child and adolescent psychiatry board certification, and 2 held consult-liaison psychiatry board certification. There was slight agreement between raters (W=0.110, $p< 0.001$).

***3.5. Construct Scores, Rated by Psychiatrists***

Construct scores, as rated by psychiatrists, are summarized in Table 2.

- Accuracy: MIND was rated as accurate 71.6% of the time and had a mean accuracy score of 2.57 (SD=0.74), while ChatGPT was rated as accurate 78.9% of the time with a mean accuracy score of 2.68 (SD=0.66). This difference was significant (p=0.021), but the effect size was negligible (r=0.073).

- Completeness: MIND was rated as complete 70.3% of the time and had a mean completeness score of 2.52 (SD=0.79), while ChatGPT was rated as complete 56.3% of the time with a mean completeness score of 2.22 (SD=0.93). This was a significant difference (p<0.001), with a small effect size (r=0.160).
- Safety: MIND was rated as safe 83.9% of the time and had a mean safety score of 2.77 (SD=0.57), while ChatGPT was rated as safe 83.4% of the time with a mean safety score of 2.78 (SD=0.52). This difference was not significant (p=0.955, r=0.002).

### *3.6 Referral Appropriateness, Rated by Psychiatrists*

Psychiatrists overwhelmingly responded that the 38 questions needed to be discussed with the patient's doctor (84.5%). There was no significant correlation found between MIND (rho=0.141, p=0.398) or ChatGPT's (rho=0.234, p=0.158) referral and psychiatrists' opinion about whether referral was indicated.

### *3.7 Preferred Responses, Rated by Psychiatrists*

The majority of psychiatrists preferred the responses generated by ChatGPT (57.6%) compared to MIND (42.4%; p=0.003). When dividing raters by experience level, early career psychiatrists (N=4) had an even greater preference for ChatGPT (71.7%) over MIND (28.3%; p<0.001); while middle career psychiatrists (N=4) had an equal preference between MIND (50.7%) and ChatGPT (49.3%; p=0.935); and late career psychiatrists (N=2) had a slight non-significant preference for MIND (53.9%) over ChatGPT (46.1%; p=0.567)

## 4. Discussion

We designed a domain-specific, fine-tuned LLM ("MIND") trained exclusively on publicly-available patient education resources. MIND was able to answer many questions about the psychiatric medication escitalopram, while appropriately deferring other questions that were not answered by its training corpus. When rated by automated computer analysis, MIND demonstrated superiority over both ChatGPT and OpenEvidence in multiple areas relevant to patient education, namely accuracy, clarity, completeness, nuance, safety, and referral appropriateness. When MIND and ChatGPT were evaluated by board-licensed psychiatrists, MIND had a statistically significant and clinically meaningful advantage over ChatGPT in providing complete responses, though there were no meaningful differences in accuracy or safety of responses. However, ChatGPT's responses were preferred the majority of the time, especially by early career psychiatrists.

Interestingly, ChatGPT tended to give shorter but less-complete responses, which were more often labeled accurate; while MIND provided longer and more-complete responses, which were labeled inaccurate more often than ChatGPT. This may demonstrate a necessary tradeoff between accuracy and completeness in answering patient questions: as answers become more detailed, they are less likely to be viewed as fully accurate (Buckland & Gey, 1994). The difference in preferences between less-experienced psychiatrists and more-experienced psychiatrists is also noteworthy. This may reflect generational differences in how younger and older psychiatrists prefer to communicate with their patients, with older physicians being more detail-oriented (Chang et al., 2019). It may also reflect differences in familiarity with the response styles of LLMs such as ChatGPT, since younger physicians are more likely to be using artificial intelligence (Doximity, 2026).

It is notable that MIND, which drew its knowledge base exclusively from resources designed specifically for patient education, provided responses at a higher grade level than ChatGPT (approximately 11th grade for MIND versus 9th grade for ChatGPT), and higher than the American Medical Association's (AMA) recommendation of a 6th grade level (Rooney et al., 2021). However, this reflects a shortcoming of the training dataset rather than our model, and demonstrates the more pervasive problem that patient education resources from medical organizations are often written at a level higher than the AMA's recommendation (Rooney et al., 2021). In fact, the average Flesch-Kincaid grade level of the resources used to train our model was 10.2 (SD=1.9). (Flesch-Kincaid scores of resources are included in the supplement.)

One strength of MIND was its ability to answer vignette-style questions encompassing multiple pieces of patient data. For example, one such question was: "*My father is a 65 year old male with major depressive disorder. He also has mild Alzheimer's disease, coronary artery disease, and high cholesterol, and takes donepezil, aspirin, and atorvastatin. Is it safe for my dad to take escitalopram for his depression?*"; to which MIND responded in a nuanced yet concise manner: "*Escitalopram can be used in a 65‑year‑old with depression, even if he also has mild Alzheimer's, heart disease and high cholesterol. There's no specific contraindication with donepezil, aspirin or atorvastatin, but the drug can increase bleeding risk when taken with aspirin and can lower sodium (hyponatremia) especially in older adults. A doctor should start with a low dose, watch his blood‑work and watch for any unusual bruising or dizziness. If he's not on a mono‑amine oxidase inhibitor (MAOI) and has no other serotonin‑boosting meds, escitalopram is generally considered safe, but it should be started and monitored by his physician.*" We expect this to be one of the major advantages of an LLM over traditional patient education resources, which cannot incorporate context or provide individualized responses to questions.

### *4.1 Limitations*

Our LLM does have inherent limitations. For example, questions involving drug interactions were generally not answerable by MIND. This problem could be addressed by training MIND on a wider corpus of drug interaction data, such as additional FDA labels. However, in most cases, questions involving drug interactions are highly context-dependent and should be answered by a clinician, so it is appropriate that MIND, designed to be conservative and safe, would defer these questions to clinician review.

Our study design has limitations, as well. For one, our study was not powered with enough psychiatrist raters to adequately determine the influence of psychiatrist experience and bias in how they evaluated LLM responses, although there is a signal that experience moderates these evaluations. We also did not include a control response (e.g., a response generated by a clinician) or blind the raters in a way that would limit bias against the use of LLMs in clinical practice (Rony et al., 2024). Notably, recent data suggests that when physicians are blinded to whether clinical information is generated by an LLM, there is no significant difference in preference for a human response versus an LLM response (Ben Shitrit et al., 2026). Finally, LLM responses for our study were generated in mid-2025, and both ChatGPT and OpenEvidence have updated their models since then and may generate different responses at present. This reflects the difficulty of scientific inquiry in keeping up with the rapid pace of technology companies. However, this also emphasizes the importance of developing a model like MIND, which can be more carefully controlled and safely deployed compared to LLMs such as ChatGPT, which are subject to the shifting priorities of their for-profit enterprises.

### *4.2 Future Directions*

Future directions include:

- Fine-tuning the model to provide responses at the 6th grade readability level, and ensure accurate responses while maintaining completeness and nuance.
- Adding additional resources to the training corpus to increase the breadth of questions the model can respond to, including about other psychiatric medications.
- Expanding the sample of psychiatrist raters to include clinicians with more varied experience to enhance generalizability and increase statistical power.
- Obtaining qualitative feedback from psychiatrist raters about their preferences for particular LLM responses, and their beliefs about using LLMs for patient education.

### *4.3 Conclusions*

Overall, our domain-specific, fine-tuned LLM generally provided accurate, complete, and safe responses to patient questions about a psychiatric medication, but requires further optimization to ensure its safety for a broad range of questions. We hope this study serves as a template to build from in designing high-fidelity LLM systems to enhance patient education and care in psychiatry.

**Table 1. Construct Scores, Rated by Rubric**

| | | | | Omnibus Test | Post Hoc Pairwise Comparisons | | |
|---|---|---|---|---|---|---|---|
| | MIND | ChatGPT | Open-Evidence | MIND vs ChatGPT vs Open-Evidence | MIND vs ChatGPT | MIND vs Open-Evidence | ChatGPT vs Open-Evidence |
| Statistical Test | Mean (Standard Deviation) | | | Kruskal-Wallis | Dwass-Steel-Critchlow-Fligner | | |
| **Accuracy** | 3.97 (0.16) | 2.97 (0.16) | 3.61 (0.68) | $X^2 = 68.9$ $p < 0.001$ $\mathcal{E}^2 = 0.610$ | W = 11.86 $p < 0.001$ | W = 4.45 $p = 0.005$ | W = -7.45 $p < 0.001$ |
| **Clarity** | 3.84 (0.37) | 3.13 (0.34) | 3.21 (0.53) | $X^2 = 43.3$ $p < 0.001$ $\mathcal{E}^2 = 0.384$ | W = 8.71 $p < 0.001$ | W = 7.16 $p < 0.001$ | W = -1.26 $p = 0.647$ |
| **Complet-eness** | 3.95 (0.23) | 3.03 (0.16) | 3.53 (0.65) | $X^2 = 60.0$ $p < 0.001$ $\mathcal{E}^2 = 0.531$ | W = 11.28 $p < 0.001$ | W = 5.06 $p < 0.001$ | W = -6.36 $p < 0.001$ |
| **Nuance** | 2.74 (0.50) | 1.89 (0.39) | 2.66 (0.67) | $X^2 = 42.0$ $p < 0.001$ $\mathcal{E}^2 = 0.371$ | W = 8.72 $p < 0.001$ | W = 0.90 $p = 0.802$ | W = -7.41 $p < 0.001$ |
| **Safety** | 3.84 (0.37) | 3.13 (0.34) | 3.55 (0.83) | $X^2 = 37.9$ $p < 0.001$ $\mathcal{E}^2 = 0.336$ | W = 8.71 $p < 0.001$ | W = 2.17 $p < 0.277$ | W = -5.86 $p < 0.001$ |
| **Referral Appropri-ateness** | 4.21 (0.47) | 3.61 (0.55) | 3.61 (1.15) | $X^2 = 19.2$ $p < 0.001$ $\mathcal{E}^2 = 0.170$ | W = -6.37 $p < 0.001$ | W = -3.63 $p = 0.028$ | W = 2.21 $p = 0.291$ |

Key. $X^2$ = Kruskal-Wallis chi-squared test statistic, p = p-value, $\mathcal{E}^2$ = Kruskal-Wallis effect size, W = Wilcoxon rank-sum test statistic.

**Table 2. Construct Scores, Rated by Psychiatrists**

| | MIND | ChatGPT | MIND vs ChatGPT | |
|---|---|---|---|---|
| **Statistical Test** | **Mean (Standard Deviation)** | | **Significance: Mann Whitney U** | **Effect Size: Rank Biserial Correlation** |
| **Accuracy** | 2.57 (0.74) | 2.68 (0.66) | p = 0.021 | 0.073 (negligible) |
| **Completeness** | 2.52 (0.79) | 2.22 (0.93) | p < 0.001 | 0.160 (small) |
| **Safety** | 2.77 (0.57) | 2.78 (0.52) | p = 0.955 | 0.002 (negligible) |

Key. p = p-value.

**Supplement 1: Patient Education Resources**

- American Academy of Child and Adolescent Psychiatry (AACAP.org):
  - Anxiety Disorders: Parents' Medication Guide: https://www.aacap.org/App_Themes/AACAP/docs/resource_centers/resources/med_guides/anxiety-parents-medication-guide.pdf
    - Flesch-Kincaid Readability Score / Grade Level: 47.9 / 10.7
  - Depression: Parents' Medication Guide: https://www.aacap.org/App_Themes/AACAP/docs/resource_centers/resources/med_guides/DepressionGuide-web.pdf
    - Flesch-Kincaid Readability Score / Grade Level: 45.4 / 10.9
- American Academy of Pediatrics (HealthyChildren.org):
  - How is Depression Treated in Children & Teens?: https://www.healthychildren.org/English/health-issues/conditions/emotional-problems/Pages/depression-in-children-and-teens-treatment-options.aspx
    - Flesch-Kincaid Readability Score / Grade Level: 60.1 / 8.4
- American Psychiatric Association (Psychiatry.org):
  - What are Anxiety Disorders?: https://www.psychiatry.org/patients-families/anxiety-disorders/what-are-anxiety-disorders
    - Flesch-Kincaid Readability Score / Grade Level: 43.1 / 11.2
  - What is Depression?: https://www.psychiatry.org/patients-families/depression/what-is-depression
    - Flesch-Kincaid Readability Score / Grade Level: 49.3 / 10.1
- FDA Medication Guide (AccessData.FDA.gov):
  - Escitalopram (Lexapro): https://www.accessdata.fda.gov/drugsatfda_docs/label/2023/021323s057,021365s040lbl.pdf
    - Flesch-Kincaid Readability Score / Grade Level: 29.8 / 13.9
- National Alliance on Mental Illness (NAMI.org):
  - Escitalopram (Lexapro): https://www.nami.org/about-mental-illness/treatments/mental-health-medications/types-of-medication/escitalopram-lexapro/
    - Flesch-Kincaid Readability Score / Grade Level: 48.7 / 10.3
- National Library of Medicine (MedlinePlus.gov):
  - Escitalopram: https://medlineplus.gov/druginfo/meds/a603005.html
    - Flesch-Kincaid Readability Score / Grade Level: 50.1 / 10.1
- UpToDate (UpToDate.com):
  - Patient education: Commonly used medicines for depression (The Basics): https://www.uptodate.com/contents/commonly-used-medicines-for-depression-the-basics

- Flesch-Kincaid Readability Score / Grade Level: 74.3 / 6.1
    - Patient education: Depression treatment options for adults (Beyond the Basics): https://www.uptodate.com/contents/depression-treatment-options-for-adults-beyond-the-basics
        - Flesch-Kincaid Readability Score / Grade Level: 46.5 / 10.5
    - Patient education: Depression treatment options for children and adolescents (Beyond the Basics): https://www.uptodate.com/contents/depression-treatment-options-for-children-and-adolescents-beyond-the-basics
        - Flesch-Kincaid Readability Score / Grade Level: 52.4 / 10.3

## Supplement 2: Questions & Responses

| | Responses | | |
|---|---|---|---|
| **Questions** | **MIND** | **ChatGPT** | **OpenEvidence** |
| **5 questions on medication mechanisms.** | | | |
| How does escitalopram work? | Escitalopram is a selective serotonin-reuptake inhibitor (SSRI). It blocks the protein that normally pulls serotonin (the “feel-good” chemical) back into brain nerve cells, so more serotonin stays in the space between cells and can keep signaling. Because it is the S-enantiomer of citalopram, it is about 100 times more potent at this task and has little effect on other neurotransmitters (dopamine, norepinephrine) or other brain receptors, which reduces many side-effects. The net result is increased serotonergic activity that helps lift mood and ease anxiety. | Escitalopram works by increasing the amount of serotonin, a brain chemical that helps regulate mood. It does this by blocking serotonin from being reabsorbed too quickly by brain cells, which keeps more of it available. This can help improve symptoms of depression and anxiety over time. | Escitalopram works by increasing the amount of serotonin, a chemical messenger in the brain that helps regulate mood. It does this by blocking the reuptake (reabsorption) of serotonin into nerve cells, making more serotonin available to improve communication between brain cells. This effect is thought to help relieve symptoms of depression and anxiety. |
| How does escitalopram treat depression? | Escitalopram works by boosting the amount of serotonin—a chemical messenger that helps regulate mood—in the brain. It is a highly selective serotonin-reuptake inhibitor (SSRI), meaning it blocks the protein that normally pulls serotonin back into nerve cells, so more stays available to travel between cells. This increased serotonin activity helps lift the low mood, improve energy and concentration, and reduce other depressive symptoms. The drug reaches steady levels after about a week of once-daily dosing, and because it hardly affects other neurotransmitters, it tends to have fewer side-effects than less selective antidepressants. | Escitalopram treats depression by boosting serotonin levels in the brain, which helps improve communication between brain cells. Over time, this steadier serotonin activity can lift mood, reduce negative thoughts, and ease other symptoms of depression like low energy, poor sleep, and loss of interest. | Escitalopram treats depression by increasing serotonin levels in the brain, which helps improve mood. It does this by blocking the reuptake of serotonin, making more of this chemical available for nerve cells to use. Clinical trials show that escitalopram, taken once daily at doses of 10–20 mg, leads to greater improvement in depression symptoms compared to placebo and helps prevent relapse in people who respond to treatment. |
| Will escitalopram affect my brain? | Yes. Escitalopram is an antidepressant that works directly in the brain. It blocks the re-uptake of serotonin, a chemical messenger that helps regulate mood, anxiety, and other emotions, so more serotonin stays available to the brain’s nerve cells. This increase can lift depressed mood, reduce anxiety, and improve related symptoms. | Yes, escitalopram affects your brain by changing how serotonin works, which can improve mood and reduce anxiety. These changes are usually helpful, but some people may notice side effects like headaches, trouble sleeping, or feeling more tired. The brain generally adjusts over a few weeks. | Escitalopram affects the brain by increasing serotonin activity in the central nervous system. It works by selectively inhibiting the reuptake of serotonin (5-HT) into neurons, which raises serotonin levels available for neurotransmission. This potentiation of serotonergic activity is linked to improved mood and reduced symptoms of |
| | Most people tolerate it well, but it can cause side-effects (e.g., sleep changes, nausea, rare serotonin-syndrome or bleeding risk) and should be taken under a doctor's supervision. | | depression and anxiety. Escitalopram has minimal effects on other neurotransmitters, such as norepinephrine and dopamine, and does not significantly interact with other receptor types. |
| Will escitalopram change my personality? | X I couldn't find relevant information in the documents. | Escitalopram doesn't change your personality. Instead, it helps ease symptoms of depression or anxiety so your natural personality can come through more clearly. Some people feel calmer, more balanced, or more like themselves once their symptoms improve. | Escitalopram does not fundamentally change a person's personality. Clinical trials in both healthy individuals and patients with depression show that escitalopram may lead to small, favorable shifts in certain personality traits—such as reduced neuroticism and increased openness or agreeableness—but these changes are generally modest and are consistent with improved mental health rather than a transformation of core personality features. |
| Will escitalopram affect my child's development? | Escitalopram can be used safely in most children, but it does require careful monitoring. In clinical studies, children and teens on an SSRI have a slightly higher chance of suicidal thoughts and may lose appetite or weight, so doctors watch mood and growth closely. Animal studies show that very high doses (far above the normal pediatric dose) can delay puberty, affect learning or memory, and slow growth, but these effects are not seen at the recommended dose. If a pregnant or nursing mother takes escitalopram, the baby may be exposed through the placenta or breast-milk; rare reports describe excess sleepiness, poor feeding or slower weight gain, and a small increased risk of a newborn lung problem (PPHN). Overall, at the prescribed dose escitalopram is not known to cause lasting developmental problems, but your doctor should regularly check your child's mood, weight, and overall health. | Escitalopram is generally considered safe for teens but is less often used in younger children. Research suggests it does not harm long-term brain development, but doctors carefully weigh benefits and risks. Close monitoring helps ensure your child's mood improves without unwanted side effects. | Escitalopram can affect a child's development. In children and adolescents, decreased appetite and weight loss have been observed, so regular monitoring of weight and growth is recommended during treatment with escitalopram or other SSRIs. The safety and effectiveness of escitalopram are established for major depressive disorder in patients 12 years and older, and for generalized anxiety disorder in patients 7 years and older; use in younger children is not established. Animal studies show possible delays in sexual maturation and reversible learning and memory disruption at high doses, but these effects have not been confirmed in humans at therapeutic doses. |
| **5 questions on treatment indications.** | | | |
| Why do people take escitalopram? | People take escitalopram because it is an antidepressant that helps balance serotonin in the brain. It is officially approved to treat | People take escitalopram to help with depression and anxiety. It works by balancing brain chemicals that affect mood, | People take escitalopram primarily for the treatment of major depressive disorder (MDD) and generalized anxiety |
| | major depressive disorder (MDD) and generalized anxiety disorder (GAD), reducing symptoms like persistent sadness, loss of interest, excessive worry, restlessness, and trouble sleeping. Doctors also prescribe it "off-label" for conditions such as obsessive-compulsive disorder, panic disorder, PTSD, social phobia, premenstrual dysphoric disorder, certain eating disorders, and even menopause-related hot flashes. In short, it's used to improve mood and lessen anxiety-related symptoms. | which can ease sadness, worry, tension, or panic. Many people use it when these symptoms interfere with daily life, relationships, or work. | disorder (GAD). It is prescribed for adults with these conditions, as well as for pediatric patients aged 12 years and older for MDD, and 7 years and older for GAD. The main reasons for prescribing escitalopram are to alleviate symptoms of depression—such as persistent sadness, loss of interest, and impaired daily functioning—and to reduce excessive anxiety and worry associated with GAD. Efficacy for these indications has been demonstrated in multiple placebo-controlled clinical trials. |
| Does escitalopram treat depression? | Yes. Escitalopram (brand name Lexapro) is an antidepressant that's officially approved to treat major depressive disorder (MDD). It works by increasing serotonin activity in the brain, which helps lift mood, improve energy, and reduce the typical symptoms of depression. It's also used for anxiety, but its primary, FDA-approved indication is for depression. | Yes, escitalopram is an antidepressant that treats depression. It works by increasing serotonin in the brain, which can improve mood, energy, sleep, and overall outlook over time. Many people start feeling better after several weeks of regular use. | Escitalopram treats depression by increasing serotonin activity in the central nervous system, which helps improve mood and reduce depressive symptoms. It is indicated for major depressive disorder in adults and in pediatric patients aged 12 years and older. The recommended starting dose is 10 mg once daily, which may be increased to a maximum of 20 mg once daily, taken orally in the morning or evening, with or without food. Clinical trials demonstrate that escitalopram leads to greater improvement in depression scores and a longer time to relapse compared to placebo. |
| Does escitalopram treat anxiety? | Yes. Escitalopram (Lexapro) is an FDA-approved antidepressant that is also indicated for generalized anxiety disorder (GAD). In addition to its use for major depression, doctors sometimes prescribe it off-label for other anxiety-related conditions such as panic disorder, social phobia, PTSD, and obsessive-compulsive disorder. It works by selectively increasing serotonin levels in the brain, which helps reduce excessive worry and the physical symptoms of anxiety. | Yes, escitalopram also treats anxiety. By boosting serotonin in the brain, it can reduce constant worry, tension, and physical symptoms like restlessness or a racing heart. Many people feel calmer and more in control after taking it regularly for a few weeks. | Escitalopram does treat anxiety. It is specifically indicated for the treatment of generalized anxiety disorder (GAD) in adults and in children aged 7 years and older. The recommended starting dose for both adults and pediatric patients is 10 mg once daily, which may be increased to a maximum of 20 mg once daily based on clinical response and tolerability. Efficacy in reducing anxiety symptoms has been demonstrated in multiple placebo-controlled clinical trials for both age groups. Regular monitoring of weight and growth is advised in children and adolescents due to possible decreased appetite and weight loss. |
| Does escitalopram treat trauma? | Yes. While escitalopram is officially approved for major | Escitalopram is not a specific treatment for trauma, but it can | Escitalopram is not approved to treat trauma or prevent |
| | depression and generalized anxiety, doctors sometimes prescribe it “off-label” for post-traumatic stress disorder (PTSD). That means it isn’t a primary FDA-approved indication, but clinical experience shows it can help reduce the anxiety, intrusive thoughts, and mood symptoms that often follow traumatic events. So, it can be used to treat trauma-related conditions like PTSD, though this use is not formally labeled. | help with related symptoms like depression, anxiety, or panic. For trauma itself, therapy—especially trauma-focused approaches—is usually more effective. Sometimes medication and therapy are used together for the best results. | posttraumatic stress disorder (PTSD). Some studies show that escitalopram may help certain people with PTSD feel better, but large, high-quality research does not support it as a first-choice treatment for trauma-related symptoms or for preventing PTSD after a traumatic event. Other treatments, like specific types of therapy (such as cognitive behavioral therapy), are recommended as the main approach for trauma and PTSD. If trauma symptoms are present, it is important to discuss all treatment options with a healthcare provider. |
| Does escitalopram treat headaches? | X I couldn't find relevant information in the documents. | Escitalopram isn’t meant to treat headaches, but some people notice fewer stress-related or tension headaches as their mood improves. In others, headaches can happen as a temporary side effect, especially when starting or changing the dose. | Escitalopram is not approved to treat headaches. It is indicated for major depressive disorder and generalized anxiety disorder, not for headache prevention or treatment. Headache is a common side effect of escitalopram, occurring more frequently than with placebo in clinical trials. |
| **5 questions on medication administration.** | | | |
| How should I take escitalopram? | Escitalopram is taken once a day, either in the morning or evening, with or without food. Most people start with a low dose (usually 5 mg or 10 mg) and the doctor may increase it slowly over a few weeks, up to a maximum of 20 mg daily. If you use the liquid form, measure the dose with the dosing spoon or oral syringe your pharmacy provides. Do not take other medicines that contain citalopram (Celexa) at the same time. To help you remember, use a pillbox, phone alarm, calendar, or ask a friend or family member for a reminder. If you miss a dose, take it as soon as you remember—unless it’s almost time for your next dose—then continue as usual; don’t double-dose. | Take escitalopram once a day, at the same time each day, with or without food. Swallow it with water and follow your doctor’s exact instructions. Don’t stop suddenly, even if you feel better—talk to your doctor first to avoid withdrawal symptoms. | Escitalopram should be taken by mouth once a day, in the morning or evening, with or without food. The usual starting dose for adults is 10 mg daily, which may be increased to a maximum of 20 mg daily if needed. Do not stop taking escitalopram suddenly; any changes should be discussed with a healthcare provider to avoid withdrawal symptoms. Always follow the instructions given by your doctor or pharmacist. |
| What happens if I miss a dose of escitalopram? | If you forget to take a dose of escitalopram, take it as soon as you remember — but only if it isn’t almost time for your next scheduled dose. If the next dose | If you miss a dose of escitalopram, take it as soon as you remember. If it’s almost time for your next dose, skip the missed one—don’t double up. | If you miss a dose of escitalopram (Lexapro), take it as soon as you remember. If it’s almost time for your next dose, just skip the missed dose—do not |
| | is near, skip the missed one and continue with your regular schedule. Do not take two doses at once or take more than prescribed. If you're unsure what to do, call your doctor or pharmacist for guidance. | Missing an occasional dose usually isn't serious, but taking it regularly helps it work best. | take two doses at once. Missing one dose is usually not harmful, but missing several doses or stopping suddenly can cause symptoms like feeling irritable, dizzy, or anxious. If you have questions or feel unwell after missing doses, contact your healthcare provider. |
| If I accidentally take an extra dose of escitalopram, what should I do? | X I couldn't find relevant information in the documents. | If you accidentally take an extra dose of escitalopram, you'll likely be fine, but you might feel dizzy, sleepy, or nauseous. Do not take more to "catch up." Call your doctor or a poison control center right away for guidance, especially if you took more than one extra dose. | If someone accidentally takes an extra dose of escitalopram, **watch for symptoms like nausea, dizziness, agitation, rapid heartbeat, or unusual behavior** and seek medical attention if these occur.[1] Serious effects such as seizures, heart rhythm changes, or serotonin syndrome are rare but possible, especially with much higher doses or if combined with other similar drugs; call Poison Help (1-800-222-1222) or go to the ER if severe symptoms develop.[1] Do not stop escitalopram suddenly; continue regular dosing unless told otherwise by a healthcare provider.[1] Most single, small extra doses are unlikely to cause harm in healthy adults, but children, elderly, or those with liver problems should contact their doctor.[1] |
| Do I have to take escitalopram every day? | Yes. Escitalopram is meant to be taken once each day, usually in the morning or evening, with or without food. Taking it every day keeps a steady level of the medicine in your body, which is needed for it to work best for depression or anxiety. If you forget a dose, take it as soon as you remember—unless it's almost time for your next dose—then continue with your regular schedule. Don't double‑up to make up for a missed pill. Consistent daily use is the safest way to get the full benefit. | Yes, you should take escitalopram every day for it to work properly. It needs to build up in your system to balance brain chemicals over time. Skipping doses can make it less effective and may cause withdrawal symptoms if stopped suddenly. | Escitalopram should be taken once every day, at about the same time, either in the morning or evening, with or without food. Daily dosing helps maintain steady levels in the body and supports its effectiveness for depression and anxiety. Stopping escitalopram suddenly can cause withdrawal symptoms, so any changes should be discussed with a healthcare provider. |
| Should I take escitalopram in the morning or at night? | X I couldn't find relevant information in the documents. | You can take escitalopram either in the morning or at night—whichever fits best with your routine. If it makes you feel more alert, take it in the morning. If it makes you sleepy, take it at night. Just take it at the same time each day. | Escitalopram (Lexapro) can be taken once daily, either in the morning or at night, depending on what works best for the individual. The medication's absorption is not affected by food, so it can be taken with or without meals. If escitalopram |
| | | | causes drowsiness, taking it at night may help; if it causes trouble sleeping, morning dosing may be better. For best results, take it at the same time each day. There is no clinical advantage to morning versus evening dosing—personal comfort and routine are most important. |
| **5 questions on dosing.** | | | |
| What dose of escitalopram should I take for depression? | For most adults with major depressive disorder, the usual starting dose of escitalopram (Lexapro) is 10 mg taken once each day. Your doctor may keep you on that dose, or if needed, may increase it to 20 mg once daily after a few weeks. Higher doses haven't shown extra benefit. If you're an older adult or have liver problems, the recommended dose stays at 10 mg daily. Always follow your prescriber's instructions and never change the dose on your own. | For depression, most adults start with 10 mg of escitalopram once daily. Your doctor may increase it to 20 mg if needed, depending on how you respond and tolerate side effects. Always follow your doctor's instructions—don't adjust the dose on your own. | The usual starting dose of escitalopram for depression is 10 mg taken once a day, with or without food, in the morning or evening. If needed, the dose can be increased to a maximum of 20 mg once a day after at least one week, depending on how well it works and if there are any side effects. Most people do well on the 10 mg dose, and higher doses do not always give extra benefit. |
| What dose of escitalopram should I take for anxiety? | For anxiety (generalized anxiety disorder), the usual starting dose of escitalopram (Lexapro) is 10 mg taken once a day. If you tolerate it well and need more relief, a doctor may raise the dose to 20 mg once daily after at least one week (in adults) or after about three weeks (in children ≥ 7 years). Older adults and people with liver problems are usually kept at the 10 mg dose. Always follow your prescriber's instructions and don't change the dose on your own. | For anxiety, most adults start with 10 mg of escitalopram once daily. If needed, your doctor may raise it to 20 mg after a few weeks. The best dose varies by person, so follow your doctor's guidance and report any side effects or changes in symptoms. | The recommended dose of escitalopram (Lexapro) for anxiety is 10 mg taken by mouth once a day, either in the morning or evening, with or without food. If needed, the dose can be increased to a maximum of 20 mg once a day after at least one week, depending on how well the medicine is working and if there are any side effects. For older adults or those with liver problems, the recommended dose is 10 mg once a day. |
| What dose of escitalopram should I take for OCD? | Escitalopram isn't officially approved for OCD, so doctors usually treat it the same way they do for depression or anxiety. The typical starting dose is 10 mg once a day, taken in the morning or evening with or without food. If needed, the dose can be raised to a maximum of 20 mg once daily—higher doses haven't shown extra benefit. Older adults or people with liver problems generally stay at 10 mg. Because OCD use is off-label, your clinician will decide the exact dose that's right for you. | Escitalopram isn't officially approved for OCD, but doctors sometimes prescribe it "off-label." Doses are usually higher—often starting at 10 mg daily and increasing up to 20 mg or more if tolerated. Your doctor will adjust based on your response and side effects. | For obsessive-compulsive disorder (OCD), the usual starting dose of escitalopram is 10 mg once daily, which can be increased to a maximum of 20 mg once daily based on response and tolerability. Higher doses may be considered in severe or treatment-resistant cases, but should only be used under close medical supervision due to increased risk of side effects. Treatment should be continued for at least 8–12 weeks to assess effectiveness, and longer for maintenance if beneficial. |
| What is the highest dose of escitalopram? | The FDA label lists the maximum recommended dose of escitalopram as 20 mg taken once daily (for adults and pediatric patients where approved). Tablets are available in 5 mg, 10 mg (scored) and 20 mg (scored) strengths, and the 20‑mg dose is the highest amount the label advises. | The highest recommended dose of escitalopram for adults is 20 mg once daily. Taking more than that doesn't usually increase benefits and can raise the risk of side effects, such as heart rhythm changes or excessive drowsiness. Always follow your doctor's instructions. | The highest recommended dose of escitalopram (Lexapro) is 20 mg taken once daily. This is the maximum dose for adults and most children, and it should not be exceeded. For elderly patients or those with liver problems, the highest recommended dose is 10 mg once daily. The medication is usually taken by mouth, with or without food, in the morning or evening. |
| When should I increase my escitalopram dose? | You should only raise the dose after you've been on a stable dose for at least one week (for adults) and you're tolerating it without troublesome side‑effects. If after that week you still don't have enough relief of your symptoms, your doctor may increase the tablet to 20 mg once daily – the highest approved amount. (For children 7 years and older, the waiting period is at least two weeks before any increase.) Always make the change under your prescriber's guidance and never jump straight to a higher dose on your own. | You should only increase your escitalopram dose if your doctor advises it. They may raise it after a few weeks if your symptoms haven't improved enough and you're tolerating the medicine well. Never adjust the dose on your own, as it can cause side effects or withdrawal. | The dose of escitalopram should be increased if symptoms of depression or anxiety have not improved after at least 1–2 weeks at the starting dose, and side effects are tolerable. The usual starting dose is 10 mg once daily, with a maximum recommended dose of 20 mg once daily for most adults; increases should be made at intervals of at least 1 week. For older adults or those with liver problems, the recommended dose is 10 mg daily and should not be increased without medical advice. Higher doses may not always provide extra benefit and can increase side effects, so dose increases should be considered only if there is insufficient response to the initial dose. |
| **5 questions on common side effects.** | | | |
| What are the common side effects of escitalopram? | Common side effects of escitalopram (Lexapro) include:<br>Nausea – feeling queasy or an upset stomach.<br>Sleep problems – trouble falling asleep or staying asleep (insomnia).<br>Fatigue or feeling tired.<br>Sexual changes – reduced libido, difficulty reaching orgasm, or delayed ejaculation.<br>These effects are seen in about 5% or more of patients and are usually the first things people notice when they start the medication. Most are mild and often improve after a couple of weeks, though sexual side effects may persist. | Common side effects of escitalopram include nausea, tiredness, trouble sleeping, dry mouth, and sweating. Some people also feel dizzy, have headaches, or notice changes in appetite or sexual desire. These effects are usually mild and improve after the first few weeks. | Common side effects of escitalopram include nausea, trouble sleeping, tiredness, sleepiness, sweating, dry mouth, headache, dizziness, diarrhea, and sexual problems such as delayed ejaculation or reduced sexual desire. These effects were seen in clinical trials, with nausea, insomnia, fatigue, and sexual dysfunction occurring in at least 5% of patients and more often than with placebo. Most side effects are mild and often improve over time. |
| What are all of the side effects of escitalopram? | X I couldn't find relevant information in the documents. | Escitalopram can cause mild to serious side effects. Common ones include nausea, fatigue, dry | Common side effects of escitalopram include nausea, trouble sleeping, tiredness, |
| | | mouth, sweating, dizziness, headache, and trouble sleeping. Some people notice changes in appetite, weight, or sexual function. Rare but serious effects include serotonin syndrome, low sodium levels, heart rhythm changes, or suicidal thoughts—especially in young people. | sleepiness, sweating, dry mouth, dizziness, headache, diarrhea, constipation, and sexual problems (such as delayed ejaculation, reduced sex drive, or trouble having an orgasm). Less common effects are increased appetite, weight gain, flu-like symptoms, and menstrual changes. Rare but serious risks include suicidal thoughts (especially in young people), serotonin syndrome (dangerous high levels of serotonin), seizures, mania, low sodium, increased bleeding, and vision changes. Some people may experience allergic reactions, muscle pain, or urinary problems. |
| What are the sexual side effects of escitalopram? | Escitalopram (Lexapro) can affect sexual function in both men and women. Common problems include:<br>Reduced sex drive (lower libido)<br>Ejaculation issues in men – mainly delayed or reduced ejaculate, and occasional impotence<br>Delayed or absent orgasm (anorgasmia) in both sexes<br>Priapism (a painful, prolonged erection) – very rare<br>These changes often start early in treatment and may not improve over time. If you notice any of these effects, talk to your doctor; they can suggest ways to manage them. | Sexual side effects of escitalopram can include lower sex drive, delayed orgasm, or trouble getting or keeping an erection. These effects are common with many antidepressants and may improve over time or with a change in dose or medication. Talk to your doctor if they bother you. | Escitalopram (Lexapro) can cause sexual side effects in both men and women. In men, these may include delayed ejaculation, decreased sexual desire, and erectile dysfunction. In women, decreased sexual desire and difficulty or inability to reach orgasm can occur. In clinical trials, up to 14% of men reported ejaculation problems and up to 7% of adults reported reduced libido, with these rates higher than placebo. These effects may be underreported, so it is important to discuss any changes in sexual function with a healthcare provider. |
| Will escitalopram affect my sex drive? | Yes. Escitalopram (Lexapro) can lower sexual desire. In clinical trials about 6 % of men and 3 % of women reported a decreased libido, and the label notes that both men and women may experience reduced sex drive, delayed orgasm or ejaculatory problems. Not everyone is affected, but the change can persist while you're taking the drug. Talk to your prescriber if you notice any drop in desire—they can adjust the dose, switch medicines, or suggest strategies to manage the side‑effect. | Yes, escitalopram can lower sex drive for some people. It may also make it harder to have an orgasm or erection. These effects vary from person to person and sometimes improve over time. If they persist, your doctor can suggest ways to manage them. | Escitalopram (Lexapro) can sometimes lower your sex drive or cause other sexual side effects, like trouble with orgasm or ejaculation. These changes happen in some people, but not everyone. Both men and women can be affected. If you notice any changes in your sexual health while taking this medicine, talk to your doctor—there are ways to help manage these side effects. |
| Will I gain weight if I take escitalopram? | X I couldn't find relevant information in the documents. | Some people gain a small amount of weight while taking escitalopram, but others don't notice any change. Weight gain | Escitalopram (Lexapro) is not associated with significant weight gain in clinical trials. Studies comparing escitalopram |
| | | can happen because mood and appetite improve, not just from the medicine itself. Regular exercise and healthy eating can help manage this effect. | to placebo found no important differences in body weight between groups.[1] However, weight gain has been reported rarely after the drug was approved and used more widely, but these cases are uncommon and not clearly proven to be caused by the medication.[1] Most people taking escitalopram do not experience noticeable changes in weight. |
| **5 questions on rare side effects.** | | | |
| Does escitalopram make you suicidal? | Escitalopram does not deliberately “make you want to kill yourself,” but it can raise the chance of suicidal thoughts or urges in some people, especially when they first start the drug or when it’s taken by children, teens, and young adults. The FDA warning and the patient‑info sheets say doctors should watch patients closely during the first few weeks for any worsening mood, anxiety, restlessness, or new suicidal ideas. In most adults the overall risk is low, and the medication’s benefit in treating depression or anxiety usually outweighs this risk when it’s used under medical supervision. | Escitalopram does not directly make most people suicidal, but like other antidepressants, it can increase suicidal thoughts in some young people, especially in the first few weeks of treatment or after a dose change. Close monitoring and talking to your doctor about any mood changes is very important. | Escitalopram can increase the risk of suicidal thoughts and behaviors, especially in children, adolescents, and young adults under age 25. This risk is highest during the first few months of treatment or when the dose is changed. In clinical trials, there were 14 additional cases per 1,000 treated patients under 18, and 5 additional cases per 1,000 in those aged 18–24, compared to placebo. Patients should be closely monitored for worsening mood or new suicidal thoughts, particularly early in treatment. |
| Does escitalopram cause serotonin syndrome? | Yes. Escitalopram (Lexapro) is an SSRI, and like other drugs in this class it can trigger serotonin syndrome—a potentially life‑threatening reaction when there is too much serotonin in the brain. The risk is highest if escitalopram is taken together with other serotonergic medicines (e.g., other antidepressants, triptans, certain pain drugs, linezolid, IV methylene‑blue) or with drugs that block serotonin breakdown (MAO inhibitors). Even on its own, rare cases have been reported, so any sudden symptoms such as agitation, fever, rapid heartbeat, tremor, or diarrhea should be evaluated promptly. | Yes, but it’s rare. Serotonin syndrome can happen if escitalopram is taken with other medicines that raise serotonin levels. Symptoms include agitation, sweating, tremors, fast heartbeat, or confusion. It’s a medical emergency—get help right away if you notice these signs. | Escitalopram (Lexapro) can cause serotonin syndrome, but this is rare when taken alone and is much more likely if combined with other drugs that increase serotonin, such as certain antidepressants, migraine medications, or MAOIs. Symptoms can include confusion, agitation, fever, sweating, muscle stiffness, and diarrhea. Patients should be monitored for these symptoms, especially when starting or increasing the dose, or when used with other serotonergic drugs. |
| How do I know if I am having serotonin syndrome? | X I couldn't find relevant information in the documents. | Signs of serotonin syndrome include restlessness, confusion, sweating, shaking, muscle stiffness, fever, fast heartbeat, or | Serotonin syndrome can be recognized by a combination of symptoms that often include feeling confused or agitated, |
| | | diarrhea. Symptoms can appear within hours of a dose change or mixing medicines. If you suspect serotonin syndrome, get emergency medical help right away. | having a fast heartbeat, sweating a lot, and muscle twitching or stiffness. People may also notice tremors, shivering, or jerky movements, especially in the legs. These symptoms usually happen after starting or increasing the dose of a medication that affects serotonin, such as certain antidepressants. If these signs appear suddenly, especially after a medication change, medical attention is needed. |
| Does escitalopram increase my risk of seizures? | Escitalopram can rarely cause seizures, especially in severe or acute reactions (such as overdose or when combined with other drugs that lower the seizure threshold). In most people it does not raise seizure risk, but if you have a history of seizures, are taking other seizure‑triggering medicines, or use high doses, you should discuss it with your doctor. | Escitalopram can very rarely increase the risk of seizures, mostly in people who already have a seizure disorder or take high doses. For most people, the risk is extremely low. Always tell your doctor if you have a history of seizures. | For most people, escitalopram (Lexapro) does not significantly increase the risk of seizures. In clinical trials, seizures were reported rarely, and people with a history of seizure disorders were excluded from these studies.[1] The drug label advises caution when prescribing escitalopram to those with a history of seizures, but for the average person without such a history, the risk is very low.[1] Seizures are more likely in cases of overdose or when escitalopram is combined with certain other medications. |
| Does escitalopram cause bleeding? | Yes. Escitalopram (Lexapro) can increase the chance of bleeding. It interferes with serotonin's role in platelet function, so people taking it may bruise more easily, have nosebleeds, gum bleeding, or even gastrointestinal bleeding—especially if they also use aspirin, ibuprofen, other NSAIDs, warfarin, or other blood‑thinners. The risk is real, though not everyone will bleed; doctors usually warn patients and monitor them when these other drugs are combined. | Escitalopram can slightly increase the risk of bleeding, especially if taken with blood thinners, aspirin, or NSAIDs like ibuprofen. This happens because serotonin helps platelets clot blood. Nosebleeds, easy bruising, or stomach bleeding are rare but possible—tell your doctor if they occur. | Escitalopram (Lexapro) can increase the risk of bleeding. This risk is higher if taken with other medications that affect blood clotting, such as aspirin, NSAIDs, warfarin, or other anticoagulants. Bleeding events can range from mild (like nosebleeds or easy bruising) to serious (such as gastrointestinal bleeding or, rarely, life-threatening hemorrhages).[1] If you are taking blood thinners or have bleeding concerns, inform your healthcare provider. |
| **5 questions on medication interactions.** | | | |
| Can I take escitalopram with ibuprofen? | X I couldn't find relevant information in the documents. | You can take escitalopram with ibuprofen occasionally, but doing so often may slightly raise the risk of stomach bleeding. If you need pain relief regularly, ask your doctor or pharmacist about safer options, such as acetaminophen (Tylenol). | Escitalopram can be taken together with ibuprofen, but combining them may increase the risk of bleeding, especially in the stomach or intestines. This is because escitalopram (an SSRI) and ibuprofen (an NSAID) both affect blood clotting. Patients should be aware of symptoms like unusual bruising, nosebleeds, or black stools and |
| | | | consult a healthcare provider if these occur. |
| Can I take escitalopram with metformin? | X I couldn't find relevant information in the documents. | Yes, escitalopram and metformin can be taken together. They don't interact in a harmful way for most people. However, if you notice changes in blood sugar, mood, or energy, let your doctor know so they can adjust your treatment if needed. | Escitalopram can generally be taken together with metformin, as there are no known direct drug interactions between them. Escitalopram is not listed as contraindicated with metformin or other common diabetes medications.[1] In fact, escitalopram may help improve blood sugar control in people with type 2 diabetes, as some studies show it can lower HbA1c and fasting blood glucose levels.[2] However, patients with severe kidney disease should consult their doctor for individual advice, since both drugs are used in people who may have kidney impairment. |
| What will happen if I take escitalopram with metoprolol? | Taking Lexapro (escitalopram) together with metoprolol raises the amount of metoprolol that gets into your bloodstream—about a 50 % higher peak level and an 80 % larger overall exposure. This can make metoprolol less cardio‑selective, but in studies healthy people did not show any meaningful changes in blood pressure or heart rate. In short, the drug interaction increases metoprolol levels, but it usually does not cause noticeable blood‑pressure or pulse problems. Still, your doctor may want to monitor you, especially if you have heart issues. | Taking escitalopram with metoprolol can slightly increase metoprolol's effects, possibly leading to a slower heart rate or lower blood pressure. Most people tolerate the combination well, but your doctor may monitor your heart rate and adjust doses if needed. | Taking escitalopram together with metoprolol can increase the amount of metoprolol in the body, which may reduce its heart-specific effects. However, studies in healthy people found no significant changes in blood pressure or heart rate when these drugs were used together. This combination is generally considered safe, but caution may be needed for people with certain heart conditions or those sensitive to metoprolol's side effects. |
| Is it safe to take escitalopram with methylphenidate? | Escitalopram can be taken with methylphenidate, but only under a doctor's close supervision. The patient‑information sheet warns that combining escitalopram with "amphetamines" (the drug class that includes methylphenidate) can raise the chance of serotonin syndrome—a rare but serious condition that causes agitation, rapid heart rate, fever, and muscle stiffness. Because of this risk, doctors usually start with low doses, watch for side‑effects, and may adjust one or both medicines. So it isn't automatically unsafe, but it isn't something you should do on your own. | Escitalopram and methylphenidate can sometimes be used together safely, especially for people with both depression or anxiety and ADHD. However, the mix can rarely increase the risk of side effects like restlessness, high blood pressure, or serotonin syndrome. Your doctor will monitor you closely if you take both. | Taking escitalopram with methylphenidate is generally considered safe for adults, with no significant increase in serious side effects compared to methylphenidate alone.[1] The combination may even lower headache risk, though escitalopram could be linked to higher blood pressure and cholesterol than some other antidepressants.[1] Both drugs can raise serotonin levels, so there is a small risk of serotonin syndrome; watch for symptoms like confusion, sweating, or tremors, especially when starting or increasing doses.[2] The FDA also advises monitoring for |
| | | | bleeding risk if other blood-thinning drugs are used. |
| Can I take escitalopram with alcohol? | X I couldn't find relevant information in the documents. | It's best to avoid alcohol while taking escitalopram. Alcohol can make side effects like drowsiness, dizziness, and poor coordination worse. It can also reduce the medication's benefit and worsen depression or anxiety symptoms. | Drinking alcohol while taking escitalopram (Lexapro) is not recommended. Although clinical trials showed escitalopram does not increase alcohol's effects on thinking or movement, the drug label advises against combining them, especially for people with depression, because alcohol can worsen symptoms and interfere with treatment. |
| **5 questions on special cases (eg, liver impairment, kidney impairment, geriatric, pregnancy).** | | | |
| Can I take escitalopram if I have a bad liver? | Yes—you can still use escitalopram if you have liver problems, but the dose is usually lowered. The FDA label recommends 10 mg once daily for patients with hepatic (liver) impairment (the same reduced dose is used for most older adults). Because a weaker liver clears the drug more slowly, a smaller amount helps keep blood levels in the safe range. Always discuss this with your doctor, who can confirm the right dose and monitor you for any side‑effects. | If you have liver problems, you can usually still take escitalopram, but your doctor may prescribe a lower dose. The liver helps break down the medicine, so people with liver disease may process it more slowly. Regular checkups help ensure it stays safe for you. | Escitalopram can be taken by people with liver disease, but the dose should be reduced to 10 mg once daily because the liver processes the drug more slowly, leading to higher levels in the body.[1] Extra caution and regular monitoring are important, as rare cases of liver injury have been reported, so checking liver function tests after starting escitalopram may be helpful, especially for those with existing liver problems.[2] There is limited information on use in severe liver disease, so close medical supervision is recommended. |
| Can I take escitalopram if I have kidney failure? | Escitalopram can be used in people whose kidneys work a little less well, but the evidence is limited when kidney function is very poor. Mild‑to‑moderate kidney problems (creatinine clearance > 20 mL/min): the drug's clearance is only about 17 % lower, and no dose change is usually needed. Severe kidney failure (creatinine clearance < 20 mL/min): the label says there's no reliable information on how the drug behaves, so doctors generally avoid it or use it only with close monitoring and possibly a lower dose. Talk with your physician; they can decide whether the benefits outweigh the unknown risks and may check blood levels or choose another antidepressant. | Yes, escitalopram can usually be taken safely in kidney failure, but your doctor may adjust the dose if your kidney function is severely reduced. Since the medicine is partly cleared through the kidneys, careful monitoring helps prevent buildup and side effects. | Escitalopram can generally be taken by people with mild to moderate kidney problems without changing the dose, but there is limited information about its safety in severe kidney failure. Doctors often lower SSRI doses for patients with poor kidney function, but some still receive standard doses, so careful monitoring for side effects is important.[1-2] Data on safety and effectiveness in advanced kidney disease are limited, and close follow-up is advised.[3] There is no specific guidance for escitalopram dosing in patients on dialysis or with end-stage kidney disease. |
| Can I take escitalopram if I am | Yes. Escitalopram (Lexapro) is | Yes, people aged 70 and older | Yes, a 70-year-old person can |
| 70 years old? | approved for use in older adults, including someone who is 70. The label recommends a starting (and usually sufficient) dose of 10 mg once daily for most elderly patients. Because the drug stays in the body longer in people over 65, doctors often begin with the lower dose and watch for side-effects such as low sodium (hyponatremia) or increased sensitivity. As long as you have no contraindicating conditions (e.g., recent MAOI use) and your doctor monitors you, taking escitalopram at 70 is considered safe. | can take escitalopram, but doctors usually start with a lower dose, such as 5 mg daily. Older adults may be more sensitive to side effects like dizziness or low sodium. Regular checkups help keep it safe and effective. | take escitalopram (Lexapro) for depression or anxiety, with a recommended starting dose of 10 mg daily for elderly patients. Most older adults tolerate escitalopram similarly to younger adults, but they may be more sensitive to side effects like low sodium levels, so regular monitoring is important. |
| Will escitalopram affect my baby if I take it while I am pregnant? | Yes. Taking escitalopram during pregnancy, especially in the second half, can increase the chance of problems for the baby. Studies have linked later-pregnancy use to a rare but serious condition called persistent pulmonary hypertension of the newborn (PPHN) and to newborn withdrawal symptoms such as breathing, temperature-control, feeding or sleep difficulties. Because stopping the medicine also raises the risk of a depression relapse, you should discuss the benefits and risks with your health-care provider before deciding. | Escitalopram can be used during pregnancy if the benefits outweigh the risks. Some studies suggest a small chance of withdrawal symptoms or breathing issues in newborns, but untreated depression can also harm both mother and baby. Always discuss risks and benefits with your doctor before deciding. | Taking escitalopram (Lexapro) during pregnancy may affect the baby, particularly if used later in pregnancy, but available data do not show an increased risk of major birth defects or miscarriage. However, neonates exposed to escitalopram or other SSRIs late in the third trimester have developed complications such as respiratory distress, temperature instability, feeding difficulties, hypotonia, tremor, irritability, and, in some cases, persistent pulmonary hypertension of the newborn (PPHN). These complications may require prolonged hospitalization, respiratory support, or tube feeding and can arise immediately upon delivery.[1]<br>The drug label also notes a less than two-fold increase in the risk of postpartum hemorrhage when SSRIs are used in the month before delivery.[1] Animal studies have shown adverse effects on embryo/fetal and postnatal development at doses greater than human therapeutic doses, but these findings have not been established in human epidemiologic studies at standard doses.[1]<br>Importantly, untreated depression during pregnancy carries its own risks, including relapse and adverse maternal and fetal outcomes, so the risks and benefits of continuing escitalopram should be carefully weighed.[1] |
| Will escitalopram affect my baby if I take it while I am breastfeeding? | Yes. Escitalopram does get into breast‑milk—infants receive about 4 % of the mother's dose (and a smaller amount of its metabolite). Some breast‑fed babies have shown excess sleepiness, restlessness, agitation, poor feeding or slower weight gain. Because the amount passed is low, many infants do fine, but doctors usually advise watching the baby for these signs and weighing them regularly. The decision to continue the medication should balance the mother's need for treatment with the benefits of breastfeeding and any possible infant effects. | Small amounts of escitalopram pass into breast milk, but most babies don't have problems from it. Some may be a little sleepy or fussy. Doctors often consider it one of the safer antidepressants for breastfeeding, but they'll monitor your baby to be sure. | Escitalopram is excreted into human breast milk, and infants exposed via breastfeeding may receive approximately 3.9% of the maternal weight-adjusted dose. According to the FDA label, there are reports of excessive sedation, restlessness, agitation, poor feeding, and poor weight gain in breastfed infants exposed to escitalopram, though these effects are not universally observed and data are limited. [1] Mothers taking escitalopram while breastfeeding should be advised to monitor their infants for these symptoms and seek medical care if they occur.[1] The developmental and health benefits of breastfeeding should be weighed against the mother's clinical need for escitalopram and any potential adverse effects on the infant.[1] There are no data on the effects of escitalopram or its metabolites on milk production.[1] |
| **5 questions on other common patient questions.** | | | |
| What are the irreversible side effects of taking escitalopram? | Escitalopram can cause a few side‑effects that may be permanent or long‑lasting, although they are rare:<br>Persistent sexual problems – delayed ejaculation, reduced libido or difficulty reaching orgasm often do not improve even after the drug is stopped.<br>Angle‑closure glaucoma – a sudden rise in eye pressure that can damage the optic nerve and lead to lasting vision loss if not treated quickly.<br>Loss of bone density – long‑term use can lower bone strength, increasing the risk of fractures that may not fully reverse.<br>Serious blood‑cell disorders (e.g., aplastic anemia, agranulocytosis) and heart rhythm abnormalities (QT‑prolongation, torsades de pointes) can cause lasting organ damage if they occur.<br>These effects are uncommon, but because they can be irreversible, any warning signs should be reported to a health‑care provider right away. | Irreversible side effects from escitalopram are extremely rare. Most side effects, like nausea or tiredness, go away after stopping the medicine. Very rarely, long-term sexual side effects (such as reduced libido) may persist after discontinuation, but this is uncommon and still being studied. | Irreversible side effects from escitalopram are extremely rare. Most side effects, such as nausea, insomnia, sexual dysfunction, and fatigue, resolve after stopping the medication. However, there have been rare reports of persistent sexual dysfunction (such as ongoing problems with orgasm or libido) even after discontinuation, and very rarely, movement disorders like tardive dyskinesia may not fully resolve. These effects are uncommon and most people do not experience permanent problems. |
| Will escitalopram affect my | X I couldn't find relevant | Escitalopram does not appear to | Taking escitalopram (Lexapro) |
| ability to have children? | information in the documents. | affect fertility in most people. Rarely, it may cause temporary sexual side effects, like reduced libido or delayed orgasm, but these usually improve after stopping or adjusting the dose. It does not harm sperm or egg health. | may make it a little harder to get pregnant, but it does not cause infertility or increase the risk of miscarriage. Some people notice changes in sexual desire or function while using this medicine, but most can still have children if they wish. If you are planning to become pregnant or have concerns, talk with your doctor about the best options for you. |
| Will I have withdrawals if I stop escitalopram? | Yes. If you stop escitalopram suddenly, you can get “withdrawal” or discontinuation symptoms such as irritability, nausea, dizziness, vomiting, nightmares, headache, or tingling‑like sensations on the skin. Even missing doses can raise the chance of these effects and of your depression or anxiety returning. To avoid them, work with your health‑care provider to taper the medication slowly rather than quitting abruptly. | Yes, stopping escitalopram suddenly can cause withdrawal symptoms such as dizziness, irritability, trouble sleeping, or “electric shock” sensations. These are usually mild and short-lived. To avoid them, your doctor will help you slowly taper the dose instead of stopping all at once. | If you stop taking escitalopram (Lexapro) suddenly, you may have withdrawal symptoms like dizziness, headache, nausea, trouble sleeping, or feeling irritable. These symptoms are usually mild and go away on their own, but can be uncomfortable for some people. To lower your risk, it’s best to slowly reduce your dose with your doctor’s help, rather than stopping all at once. About one in six people may have symptoms when stopping antidepressants, but most do not have severe problems. |
| When will I start to feel better after taking escitalopram? | X I couldn't find relevant information in the documents. | Most people start to feel some improvement after 2 to 4 weeks of taking escitalopram, but it can take up to 6 to 8 weeks for full benefits. Early changes may include better sleep or energy before mood and anxiety improve. Keep taking it as prescribed, even if progress feels slow. | It’s common to wonder when you’ll start to feel better after beginning escitalopram. Most people notice some improvement in mood or anxiety after about 2 weeks, but it can take 4 to 8 weeks, or sometimes longer, to feel the full benefit. Some side effects, like nausea or headache, may happen early but usually get better within the first couple of weeks. If you don’t feel any better after 4 weeks, let your doctor know so your treatment can be reviewed. |
| How long should I take escitalopram for? | Escitalopram is usually taken for several months to treat depression or anxiety, and many people stay on it longer if it continues to help. The drug has no known safety problems with long‑term use, but you should never stop abruptly—your doctor will taper the dose gradually to avoid withdrawal symptoms. So the exact length depends on your condition, how well you respond, and your doctor’s recommendation; some patients use it for a few months, others | Most people take escitalopram for at least 6 to 12 months after feeling better to prevent symptoms from returning. Some may need longer treatment, especially if depression or anxiety has come back before. Your doctor will help decide when it’s safe to stop. | Escitalopram is usually taken for several months, and often for at least 6–12 months after symptoms improve, to help prevent relapse of depression or anxiety. The exact duration depends on individual needs and response to treatment. When stopping escitalopram, the dose should be reduced gradually to avoid withdrawal symptoms. Always discuss any changes in medication with a healthcare provider to ensure safe and effective treatment. |
| | for years, always under medical supervision. | | |
| **5 questions with specific patient vignettes.** | | | |
| I am a 30 year old female who is experiencing a major depressive episode for the first time in my life. I have no other medical problems and do not take any other medications. What dose of escitalopram should I take to treat my major depressive episode? | For an adult (like you) with a first-time major depressive episode, the usual starting dose of escitalopram (Lexapro) is 10 mg taken once each day. Your doctor may keep you at that dose, or after about a week (and if you're tolerating it well) they might raise it to 20 mg once daily – the maximum recommended amount. Always follow your prescriber's instructions and let them know how you feel as you start the medication. | For a 30-year-old woman experiencing a first episode of major depression, the usual starting dose of escitalopram is 10 mg once daily. Most people stay at this dose, but if symptoms don't improve after several weeks, your doctor may increase it to 20 mg daily. Always follow your doctor's guidance and don't change the dose on your own. | The recommended starting dose of escitalopram for a 30-year-old woman with her first episode of major depression is 10 mg taken by mouth once daily. The dose can be taken in the morning or evening, with or without food. If needed, and if she tolerates it well, the dose may be increased to a maximum of 20 mg once daily after at least one week, but most people do well on the starting dose.[1] No extra precautions or dose adjustments are needed for healthy adults. |
| I am a 30 year old female who recently experienced a major depressive episode for the first time in my life. I have been taking escitalopram 10mg for 2 months now and my symptoms of depression have completely resolved. How long should I continue to take escitalopram? Can I stop taking it now? | Escitalopram isn't meant to be stopped the moment you feel better. After a first-time depressive episode, doctors usually keep you on the medicine for at least 6 months after your symptoms have cleared, and often longer (sometimes a year or more) to lower the chance of the depression coming back. Only your prescriber can decide the exact length based on how you've responded, any side-effects, and your personal risk factors. If you stop abruptly you may get withdrawal symptoms (dizziness, nausea, irritability, etc.), so any taper should be done under medical supervision. Talk to your clinician before changing or stopping the dose. | It's great that you're feeling better. Even after symptoms resolve, doctors usually recommend continuing escitalopram for at least 6 to 12 months after recovery to lower the risk of relapse. Stopping too soon can increase the chance of depression returning. When the time comes, your doctor can help you taper off slowly to avoid withdrawal symptoms. | For a first episode of major depressive disorder with complete symptom resolution, escitalopram should generally be continued for at least 6 months after remission before considering discontinuation, as this reduces the risk of relapse.[1-2] Stopping the medication now, after only 2 months of treatment, is not recommended.<br>Discontinuation should be gradual, with the dose tapered over 2–3 months to minimize discontinuation symptoms such as irritability, dizziness, and insomnia; abrupt cessation can lead to significant withdrawal effects.[1][3] If intolerable symptoms occur during tapering, resuming the previous dose and then tapering more slowly is advised.[3]<br>For patients with a single episode and no additional risk factors, expert consensus suggests considering discontinuation after 1 year of remission, though 6 months is the minimum recommended duration.[2][4] Longer maintenance may be warranted for those with recurrent episodes or severe initial symptoms.<br>Monitoring for warning signs of relapse and discontinuation symptoms during and after tapering is essential, with a plan for follow-up assessment.[1][3] |
| | | | The optimal duration of maintenance therapy after a first episode is not definitively established and should be individualized. |
| My father is a 65 year old male with major depressive disorder. He also has mild Alzheimer's disease, coronary artery disease, and high cholesterol, and takes donepezil, aspirin, and atorvastatin. Is it safe for my dad to take escitalopram for his depression? | Escitalopram can be used in a 65‑year‑old with depression, even if he also has mild Alzheimer's, heart disease and high cholesterol. There's no specific contraindication with donepezil, aspirin or atorvastatin, but the drug can increase bleeding risk when taken with aspirin and can lower sodium (hyponatremia) especially in older adults. A doctor should start with a low dose, watch his blood‑work and watch for any unusual bruising or dizziness. If he's not on a mono‑amine oxidase inhibitor (MAOI) and has no other serotonin‑boosting meds, escitalopram is generally considered safe, but it should be started and monitored by his physician. | Yes, escitalopram can be safe for your father, but it should be prescribed and monitored by his doctor. It generally works well in older adults and is safe with donepezil, aspirin, and atorvastatin, though there are a few cautions. Escitalopram can rarely lower sodium or slow the heart rate, so his doctor may start with a low dose (5 mg daily) and monitor for side effects or heart rhythm changes. | Escitalopram is generally considered safe and well-tolerated for depression in older adults with heart disease, and is among the preferred antidepressants for patients with comorbid cardiovascular conditions.[1-2] Studies show escitalopram does not significantly increase the risk of major heart problems compared to placebo in patients with heart disease, though regular monitoring is advised.[2] Escitalopram can increase bleeding risk when taken with aspirin, so it is important to watch for signs of unusual bleeding.[3] The recommended dose for older adults is 10 mg daily.[3] Common side effects include sexual dysfunction, low sodium, and rarely, cognitive changes, which may be more noticeable in those with Alzheimer's disease.[1][3]<br>No significant drug interactions are expected with donepezil or atorvastatin, but all medications should be reviewed regularly.[4] Because data in patients with both Alzheimer's and multiple health issues is limited, close follow-up is important. |
| I am a 35 year old male with bipolar disorder. I have been feeling very depressed. Should I take escitalopram? | Escitalopram can help depression, but in people with bipolar disorder it can sometimes trigger a switch into mania or rapid‑cycling mood swings. Because of that, doctors usually don't start an antidepressant like escitalopram alone; they first make sure you're on a mood‑stabilizing medication (such as lithium, valproate, or a similar drug) and they monitor you closely for any signs of becoming overly "high," irritable, or having less need for sleep. So, before you start escitalopram you should talk with your psychiatrist. They can decide whether it's safe for you, whether you need a mood | If you have bipolar disorder, you should not start escitalopram on your own. While it can help with depression, antidepressants like escitalopram can sometimes trigger mania or rapid mood swings if taken without a mood stabilizer. Talk to your psychiatrist first—they can decide if it's appropriate and ensure you're also on a medication that prevents mania. | Escitalopram is not generally recommended as a first-line treatment for bipolar depression, and taking it alone may not be safe. Antidepressants like escitalopram can increase the risk of switching to mania or hypomania, especially if not combined with a mood stabilizer or antipsychotic.[1-3] Guidelines recommend mood stabilizers (such as lithium or lamotrigine) and certain antipsychotics (like quetiapine or lurasidone) as safer and more effective first-line options for bipolar depression.[2][4-5]<br>If depression persists despite these treatments, escitalopram may be considered as an add-on, |
| | stabilizer added, and how to watch for early warning signs of mania. Do not start it on your own. | | but its benefit is modest and close monitoring is needed due to the risk of mood switching.[1-2][4][6] There is limited evidence supporting escitalopram's effectiveness and safety in bipolar disorder, so further research is needed. |
| I am a 45 year female who is pregnant. Is it safe for my baby if I take escitalopram? | Escitalopram (Lexapro) can cross the placenta, and using it in the later weeks of pregnancy has been linked to a higher chance of newborn breathing problems (persistent pulmonary hypertension) and withdrawal‑type symptoms after birth. However, stopping the drug can cause a relapse of depression, which also harms both mother and baby. Because the risks and benefits are closely balanced, you should discuss your situation with your health‑care provider—talk about how severe your depression is, any past medication reactions, other medicines you take, and your plans for breastfeeding. Together you can decide whether to stay on escitalopram or switch to another treatment. | Escitalopram can be used during pregnancy if the benefits outweigh the risks. Most research shows it's relatively safe, but some babies may have mild withdrawal symptoms or breathing issues after birth. Untreated depression can also harm both mother and baby, so it's important to talk with your doctor to decide what's best for you and your pregnancy. | If you take escitalopram (Lexapro) while pregnant, studies show it does not increase the risk of major birth defects for your baby. However, if taken late in pregnancy, some babies may have mild, short-term symptoms after birth, like fussiness, trouble feeding, or breathing issues, but these usually go away in a few days. Untreated depression during pregnancy can also be harmful, so it's important to talk with your doctor about what's best for you and your baby. |